\documentclass[conference]{IEEEtran}

\usepackage{fancyhdr}

\usepackage{cite}
\usepackage{amsmath,amssymb,amsfonts}
\usepackage{algorithm}
\usepackage{graphicx}
\usepackage{textcomp}
\usepackage[table, dvipsnames]{xcolor}

\def\BibTeX{{\rm B\kern-.05em{\sc i\kern-.025em b}\kern-.08em
    T\kern-.1667em\lower.7ex\hbox{E}\kern-.125emX}}

\usepackage{algpseudocode}
\usepackage{hyperref}
\usepackage{multirow}
\usepackage{ctable}

\newcommand{\comm}[1]{\textcolor[rgb]{0.0,0.0,1.0}{#1}}

\newcommand{\tweakedsim}{\raise.17ex\hbox{$\scriptstyle\mathtt{\sim}$}}

\definecolor{boxgray}{HTML}{E0E6E6}
\def\graybox{\cellcolor{boxgray}}

\begin{document}

\title{Democratizing AI: Open-source Scalable LLM Training on GPU-based Supercomputers}

\author{\IEEEauthorblockN{Siddharth Singh\IEEEauthorrefmark{2}$^{,1}$, Prajwal Singhania\IEEEauthorrefmark{2}, Aditya Ranjan\IEEEauthorrefmark{2}, John Kirchenbauer\IEEEauthorrefmark{2}, Jonas Geiping\IEEEauthorrefmark{1}, Yuxin Wen\IEEEauthorrefmark{2},\\Neel Jain\IEEEauthorrefmark{2}, Abhimanyu Hans\IEEEauthorrefmark{2}, Manli Shu\IEEEauthorrefmark{2}, Aditya Tomar\IEEEauthorrefmark{4}, Tom Goldstein\IEEEauthorrefmark{2}, Abhinav Bhatele\IEEEauthorrefmark{2}$^{,2}$}
\IEEEauthorblockA{~\\
\IEEEauthorrefmark{2}\textit{Department of Computer Science, University of Maryland, College Park, USA}\\
\IEEEauthorrefmark{1}\textit{ELLIS Institute T\"ubingen, Max Planck Institute for Intelligent Systems, T\"ubingen, Germany}\\
\IEEEauthorrefmark{4}\textit{Department of Electrical Engineering and Computer Sciences, University of California, Berkeley, USA}\\
}
E-mail: $^{1}$ssingh37@umd.edu, $^{2}$bhatele@cs.umd.edu
}

\maketitle

\begin{abstract}
Training and fine-tuning large language models (LLMs) with hundreds of billions
to trillions of parameters requires tens of thousands of GPUs, and a highly
scalable software stack. In this work, we present a novel four-dimensional
hybrid parallel algorithm implemented in a highly scalable, portable,
open-source framework called AxoNN. We describe several performance
optimizations in AxoNN to improve matrix multiply kernel performance, overlap
non-blocking collectives with computation, and performance modeling to choose
performance optimal configurations.  These have resulted in unprecedented
scaling and peak flop/s (bf16) for training of GPT-style transformer models on
Perlmutter (620.1 Petaflop/s), Frontier (1.381 Exaflop/s) and Alps (1.423
Exaflop/s).

While the abilities of LLMs improve with the number of trainable parameters,
so do privacy and copyright risks caused by
 memorization of training data, which can cause disclosure of sensitive or private information at inference time. We
highlight this side effect of scale through experiments that explore
``catastrophic memorization,'' where models are sufficiently large to memorize training data in a single pass, and present an approach to prevent it.
As part of this study, we demonstrate fine-tuning of a 405-billion parameter
LLM using AxoNN on Frontier.

\end{abstract}


\section{Justification for ACM Gordon Bell Prize}
\label{sec:justification}
% indicate what implementation or performance “high watermark” was achieved
% (rather than the science that was enabled)

A novel four-dimensional hybrid parallel approach to scale neural network
training to tens of thousands of AMD and NVIDIA GPUs. Time-to-solution for
80-billion parameter GPT-style transformer models was reduced by
56.0$\times$, due to kernel tuning, aggressive overlap, and
optimization of collective communication. Unprecedented performance of 1.423
Exaflop/s on 6,144 NVIDIA H100 GPUs, 1.381 Exaflop/s on 32,768 AMD MI250X GCDs,
and 620.1 Petaflop/s on 4,096 NVIDIA A100 GPUs in half-precision (bf16).

\section{Performance Attributes}
\label{sec:attributes}
% (use a table listing each attribute title and value in a separate row)

% Category of achievement (1+ of:  scalability, time-to-solution, peak
% performance)

% Type of method used (1 of:  explicit, implicit, both explicit and implicit,
% semi-implicit, n/a)

% Results reported on the basis of (1 of:  whole application including I/O;
% whole application except I/O; kernel only; other [specify])

% Precision reported (1 of:  single precision, double precision, mixed
% precision)

% System scale (1 of:  results measured on full-scale system, projected from
% results of smaller system, other [specify])

% Measurement mechanism (1 of:  timers, FLOP count, static analysis tool,
% performance modeling, other [specify] )

\begin{table}[h]
  \centering
  \begin{tabular}{p{1.4in}p{1.7in}} \toprule
  \graybox Category of achievement & \graybox peak performance, scalability, time-to-solution \\
  Type of method used & n/a \\
  \graybox Results reported on the basis of & \graybox whole application including input \\
  Precision reported & mixed-precision \\
  \graybox System scale & \graybox results measured on full-scale system \\
  Measurement mechanism & timers and FLOP count \\ \bottomrule
  \end{tabular}
  \label{tab:just}
\end{table}

\section{Overview of the Problem}
\label{sec:overview}
% description of the problem and its importance, in terms understandable to a non-specialist

% \fix{1 page max}

The field of generative artificial intelligence (AI) has taken the world by
storm. In particular, large language models (LLMs) and their chatbot interfaces
have become ubiquitous, employed by students, researchers, and professionals in
various fields on a daily basis. Modern generative AI models are built by
training extremely large neural networks, which have been shown to generalize
extremely effectively with increases in model size. This unprecedented scaling of
neural network training has been enabled by the emergence of highly efficient
GPUs, and the availability of open source training frameworks such as PyTorch,
and TensorFlow.

% Deep neural networks (DNN) are also being used in computational science, from
% climate simulations to materials modeling.
Training large neural networks that
do not fit on a single GPU with 40-96 GB of RAM requires partitioning the model
across multiple GPUs and parallelizing the matrix-matrix multiplication
operations, which are a significant fraction of the overall computation.
Scalability and parallel efficiency of DNN training is impacted by several
factors -- sustained flop/s and scalability of parallel matrix multiplication,
performance of collective communication operations over sub-communicators, and
the degree of overlap of computation with non-blocking collectives. While
classical parallel algorithms for matrix multiplication such as SUMMA and
Cannon's 2D Matrix Multiply exist, they lead to significant communication
bottlenecks when training models with hundreds of billions of parameters on
hundreds of GPUs. All of these factors make efficient parallelization of DNN
training a formidable task.
% Unlike traditional scientific computing simulations, most of the
% communication inefficiencies stem from the use of collectives like all-gather
% and reduce-scatter.

In this work, we target the challenging research problem of training models
with hundreds of billions of parameters on the fastest supercomputers with
thousands of GPUs. Traditionally, training at such scales has been restricted
to companies with large budgets and access to significant GPU resources.
However, programs such as INCITE for access to DOE supercomputers, 
Frontier and Perlmutter, and recent access to Alps at CSCS have
enabled our team to solve the research challenges in the area of parallel
training, and innovate in the area of AI/ML, by training and
fine-tuning LLMs.

There are several different challenges in ensuring scalability and a high
fraction of peak flop/s for parallel training. First, we need to ensure that
math libraries such as cuBLAS and rocBLAS are highly performant for matrix
multiplication on NVIDIA and AMD GPUs respectively. Second, we have to ensure
that both intra-node and inter-node communication of data (activations,
parameters, and gradients) is performant, and overlapped with computation as
much as possible. Third, when running on large GPU partitions with hundreds to
thousands of GPUs, the decomposition of work and its mapping to GPUs should be
near-optimal, taking communication into account.

\begin{table*}[t]
	\centering
    \caption{Comparison of large-scale LLM training studies, covering diverse
frameworks and hardware. For each study, we list the largest hardware counts
used, corresponding model \& batch size, percentage of peak flop/s, and actual
sustained flop/s.}
	\begin{tabular}{lcrrccrr} \toprule
		\textbf{Study} & \textbf{Framework} & \textbf{Model Size} & \textbf{Batch Size} & \textbf{Hardware} & \textbf{Scale} & \textbf{\% Peak} & \textbf{Petaflop/s}\\ \midrule
		SUPER~\cite{super2021jain}             & LBANN                   & 3B*   & 0.5M* & NVIDIA V100 & 1,024 GPUs & - & -\\
		KARMA~\cite{wahib2020scaling-sc}       &   KARMA                 & 17B   & 2.0M* & NVIDIA V100 & 2,048 GPUs & - & - \\
		FORGE~\cite{yin2023forge}              & GPT-NeoX                & 1.44B & 16.8M & AMD MI250X  & 2,048 GCDs & $\sim$29\%$^{\dagger}$ & $\sim$112.6$^{\dagger}$\\
		Dash et al.~\cite{dash2023optimizing}  & Megatron-DeepSpeed      & 1000B & 19.7M & AMD MI250X  & 3,072 GCDs & 31.9\%$^{\ddagger}$ & 188.0$^{\ddagger}$\\
		MT-NLG~\cite{megatron-turing-nlg-530b} & Megatron-DeepSpeed      & 530B  & 4.0M  & NVIDIA A100 & 3,360 GPUs & 36\% & 379.7\\
		Narayanan et al.~\cite{megatronlm-2}   & Megatron-LM             & 1000B & 6.3M  & NVIDIA A100 & 3,072 GPUs & 52\% & 502.0 \\
		MegaScale~\cite{jiang2024megascale}    & MegaScale               & 175B  & 12.5M & NVIDIA A100 & 12,288 GPUs& 55\%  & 2166.3 \\
		Google~\cite{gcloudscaletraining}      & Cloud TPU Multislice Training & 32B   & 417M  & TPUv5e      & 55,094 TPUs& 44.67\% & 4480.0\\
		\midrule
		\multirow{3}{*}{\bf This Work}         & \multirow{3}{*}{\textbf{AxoNN}~\cite{singh:ipdps2022}} & 40B & 16.8M & NVIDIA A100 & 4,096 GPUs & 49\% & 620.1\\
		& & 320B & 16.8M & AMD MI250X  & 32,768 GCDs & 22\% & \textbf{1381.0} \\
		& & 60B  & 16.8M & NVIDIA H100 & 6,144 GPUs  & 23\% & \textbf{1423.1} \\
		\bottomrule
		\vspace{-0.5em}\\
		\multicolumn{8}{l}{\small{* Estimated from description in paper as exact number not mentioned}}\\
		\multicolumn{8}{l}{\small{$^{\dagger}$ Estimated from plots in the paper as exact numbers not mentioned}}\\
		\multicolumn{8}{l}{\small{$^{\ddagger}$ Calculated from flop/s at lower GPU/GCD count and weak scaling efficiency}}\\
	\end{tabular}
    \label{tab:study-comparison}
\end{table*}

In order to overcome the challenges mentioned above, we have developed a
four-dimensional (4D) hybrid parallel approach that combines a
three-dimensional (3D) matrix multiplication algorithm with data parallelism to
achieve high efficiency at large GPU counts. In addition, we have improved the
performance of our implementation using several optimizations.  First, we tune
our matrix multiplication for each individual platform. Second, we aggressively
overlap computation with non-blocking collectives used in different phases of
training. Third, since the 4D algorithm requires arranging the GPUs in an
allocated job partition into a 4D virtual grid, this results in several
potential configurations, not all of which are performance optimal. Hence, we
have developed a communication model for 4D hybrid algorithms that can predict
high-performing configurations given a problem size and number of GPUs. We have implemented 
all of the aforementioned innovations in AxoNN~\cite{singh:ipdps2022, singh:ipdps2023}, our open source 
framework for large scale parallel deep learning.

We have benchmarked and optimized AxoNN on both NVIDIA and AMD GPU-based
platforms, and we present results on Perlmutter at NERSC/LBL, Frontier at
OLCF/ORNL, and Alps at CSCS. We use a range of neural network sizes with 5
billion to 320 billion parameters. AxoNN achieves an unprecedented performance
of 620.1 Petaflop/s on 4,096 NVIDIA A100 GPUs, 1.381 Exaflop/s on 32,768 MI250X
GCDs, and 1.423 Exaflop/s on 6,144 NVIDIA H100 GPUs in half-precision (bf16).

Access to a highly scalable training framework and large supercomputers have
also enabled the AI researchers on our team to study the inner workings of LLMs
at model sizes that are impossible to study otherwise. One such problem is
studying whether LLMs memorize training data and regenerate it verbatim during
inference. This has privacy risks when personally identifiable information
(PII) is memorized, and legal/copyright risks when models reproduce verbatim
copies of text without the necessary copyright and licensing information. We
present a study that explores the relationship between model size and the
memorization properties of LLMs, and demonstrate the impact of a solution to
reduce memorization.

\section{State of the Art in Parallel Training}
\label{sec:sota}
% quantitative discussion of current SoA, with emphasis on performance-related
% aspects

% \fix{1 page max}

In this section, we present the state of the art in scaling parallel training
of deep neural networks to large-scale HPC systems and data centers.

\subsection{Methods for Parallel DNN Training}
\label{sec:sota-distrain} 

Deep neural network training is typically parallelized using one of three
approaches -- data parallelism, tensor parallelism, or pipeline parallelism, or
a hybrid approach that combines some of the above.  In data parallelism, all
GPUs are assigned a full copy of the neural network, and parallelism is
achieved by sharding the input batch equally among the GPUs.  The main drawback
of data parallelism is that it requires the entire network to fit on a single
GPU. To mitigate this, sharded data parallelism has been developed, which
divides the network parameters across GPUs~\cite{sc2020zero,fsdp, wang2024zero}
and performs extra communication to gather them when needed.

Model parallelism is used to train DNNs that far exceed the memory capacity of
a single GPU, and it can be further divided into two approaches -- tensor
parallelism~\cite{megatronlm} and pipeline
parallelism~\cite{huang2019gpipe_nips, megatronlm-2}.
The former parallelizes
the computation of each layer of the neural network across several GPUs, and is
the focus of our work.  In pipeline parallelism, entire layers are assigned to
each GPU.  A popular framework for parallel training is Shoeybi et
al.'s Megatron-LM~\cite{megatronlm}, which uses a tensor parallel algorithm to
parallelize a pair of fully-connected layers.

Several frameworks combine multiple approaches to develop hybrid
methods. Narayanan et al.~\cite{megatronlm-2} extend Megatron-LM to support
hybrid parallelism by combining tensor, pipeline, and data parallelism.
Rajbhandari et al.~introduce a sharded data parallelism approach called ZeRO~\cite{sc2020zero},
which is combined with pipeline and tensor parallelism in Microsoft's training
framework, DeepSpeed~\cite{deepspeed-extreme-3d, singh:ics2023}.
% ZeRO-Infinity~\cite{zero_infinity} extends ZeRO to support training models on
% heterogeneous hardware using GPU, CPU, and NVMe memory.
% ZeRO++~\cite{wang2023zero} further extends ZeRO with communication
% optimizations to improve the performance of training large language models.
Megatron-DeepSpeed uses Megatron-LM's tensor 
parallelism.

Several other frameworks that further optimize DNN training have been proposed
in recent times. GPT-NeoX builds upon Megatron-LM and
DeepSpeed for ease of usage~\cite{gpt-neox-library}.
Wahib et al.~introduce KARMA, 
an out-of-core data parallelism framework, managing CPU-GPU data transfers to
alleviate GPU memory constraints~\cite{wahib2020scaling-sc}.
Zheng et al.~propose Alpa for 
automating neural network parallelization, optimizing communication across GPUs~\cite{alpa}.
Colossal-AI~\cite{colossalai2023unified} offers a unified interface for
distributed deep learning training.

\subsection{Large-scale Studies of Training LLMs}
\label{sec:ls-studies}

We now present recent studies on training large language models on some of the
largest GPU-based clusters.  Meta trained Llama 2~\cite{touvron2023llama} on
2000 NVIDIA A100 GPUs.  Jain et al.~\cite{super2021jain} benchmark the training
of a variant of T5-Large~\cite{t5-transformer} on 1024 NVIDIA V100 GPUs using
their proposed sub-graph parallelism technique within the LBANN
framework~\cite{essen2015lbann}.  Wahib et al.~\cite{wahib2020scaling-sc} use
KARMA to benchmark a 17B parameter model on 2048 NVIDIA V100 GPUs and report a
1.35x training speedup compared to ZeRO~\cite{sc2020zero}. Narayanan et
al.~present a weak scaling study of Megatron-LM's pipeline parallelism,
achieving 52\% of the peak NVIDIA A100 flop/s when benchmarking the training of
a 1000B parameter model on 3072 GPUs~\cite{megatronlm-2}. Shaden et
al.~\cite{megatron-turing-nlg-530b} use Megatron-LM and
DeepSpeed~\cite{deepspeed-extreme-3d} to train a 530B parameter language model
on the Selene supercomputer~\cite{selene} using 4480 NVIDIA A100 GPUs. They
achieved 113 Tflop/s per GPU with 3360 GPUs, equivalent to 36\% of the peak
performance. Ziheng et al.~\cite{jiang2024megascale} introduce MegaScale, a
production system for training LLMs at scale, achieving a 55.2\% of the peak
flop/s when benchmarking a 175B parameter model on 12,288 NVIDIA A100 GPUs.

With the emergence of AMD GPUs, several studies have focused on training large
language models on AMD systems. Yin et al.~\cite{yin2023forge} train FORGE, an
open suite of large language models for scientific computing on
Frontier~\cite{frontier2023}. In their work, the authors show scaling
performance for training FORGE on up to 2048 AMD MI250X GPUs, achieving 28\% of
the peak flop/s.  Dash et al.~\cite{dash2023optimizing} analyze efficient
distributed training strategies on Frontier for training large language models.
They achieve 31.96\% of peak when benchmarking the training of
a 1T parameter model on 1024 MI250X GPUs.

Google conducted a study on large-scale training jobs for LLMs using over
50,000 TPUv5e chips~\cite{gcloudscaletraining}. The authors achieve 44.67\% of
the peak TPUv5e performance when benchmarking a 32B parameter model on 
50,944 TPUv5e chips.  Table \ref{tab:study-comparison} provides a summary of these
studies, indicating the largest scale (in terms of the number of
GPUs/GCDs/TPUs) used and the corresponding flop/s achieved. The last set of
row in the table presents the results of this work using our open-source
training framework, AxoNN. With the exception of MegaScale, AxoNN has been
scaled to the largest number of NVIDIA GPUs. To the best of our knowledge,
AxoNN is the first framework to run on up to 32,768 AMD MI250X GCDs to achieve
a sustained flop/s performance of 1.381 Exaflop/s.

\section{Innovations Realized}
\label{sec:innovations}
% what the innovations are and how they were achieved

% \fix{2 pages max}

Training deep neural networks on a single GPU involves processing subsets of
the data called batches through the layers of a DNN in the forward pass to
compute a loss, computing the gradient of the loss in a backward pass via
backpropagation, and updating the parameters (also called ``weights'') in the
optimizer step. These three steps are repeated iteratively until all batches
have been consumed, and this entire training process is referred to as an
epoch. We now describe our novel approach to scaling the computation in the
steps described above in the context of large multi-billion parameter neural
networks on thousands of GPUs.

\subsection{A Four-Dimensional Hybrid Parallel Approach}
\label{sec:hybrid}

We have designed a hybrid parallel approach that combines data parallelism with
three-dimensional (3D) parallelization of the matrix multiplication routines.

\vspace{0.08in}
\noindent{\bf Data Parallelism:}
% When using only data parallelism, we first instantiate a full copy of the
% neural network on every GPU, and then divide the input batch into equal-sized
% {\em shards} among these GPUs.
In order to use a hybrid approach that combines data with tensor
parallelism, we organize the total number of GPUs, $G$, into a virtual 2D
grid, $G_{\mathrm{data}} \times G_{\mathrm{tensor}}$. This results in
$G_{\mathrm{data}}$ groups of $G_{\mathrm{tensor}}$ GPUs each. We use data
parallelism across the $G_{\mathrm{data}}$ groups, and tensor parallelism
within each group. Each $G_{\mathrm{data}}$ group collectively has a full copy of the neural
network and is tasked to process a unique shard of the input batch. At the end of an input batch,
all groups have to synchronize their weights by issuing all-reduces
on their gradients after every batch (this is also referred to as an
iteration).

\vspace{0.08in}
\noindent{\bf 3D Parallel Matrix Multiplication (3D PMM):}
Next, we use each GPU group, composed of $G_{\mathrm{tensor}}$ GPUs to
parallelize the work within their copy of the neural network.  This requires
distributing the matrices, and parallelizing the computation within every layer
of the neural network across several GPUs. Note that most of the computation in 
transformers is comprised of large matrix multiplications within fully connected (FC) layers.
Hence, in this section, we will focus on parallelizing FC layers with a 3D 
PMM algorithm.

We now describe how a single layer is parallelized, and the same method is
applied to all the layers in the neural network.  Each FC layer computes one
half-precision (fp16 or bf16) matrix multiplication (input activation, $I$
multiplied by the layer's weight matrix, $W$) in the forward pass and two
half-precision matrix multiplications (MMs) in the backward pass
($\frac{\partial L}{\partial O} \times W^{\top}$ and $I^{\top} \times
{\frac{\partial L}{\partial O}}$, where $L$ is the training loss, and $O$ is
the output activation.) Thus, parallelizing an FC layer requires parallelizing
these three MM operations across multiple GPUs.

% \begin{figure}[h]
%     \centering
%       \includegraphics[width=3in]{figs/fc.pdf}
%       \caption{Computation in the forward pass of a fully-connected (FC) layer with input $I$ and layer weights $W$. 
%       The output, $O$ is a matrix multiplication of $I$ and $W$. We assume $I \in \mathbb{R}^{m \times k}$, 
%       $W \in \mathbb{R}^{k \times n}$, and  $O \in \mathbb{R}^{m \times n}$. \label{fig:schematic-fc}}
% \end{figure}  

\begin{figure}[h]
    \centering
      \includegraphics[width=\columnwidth]{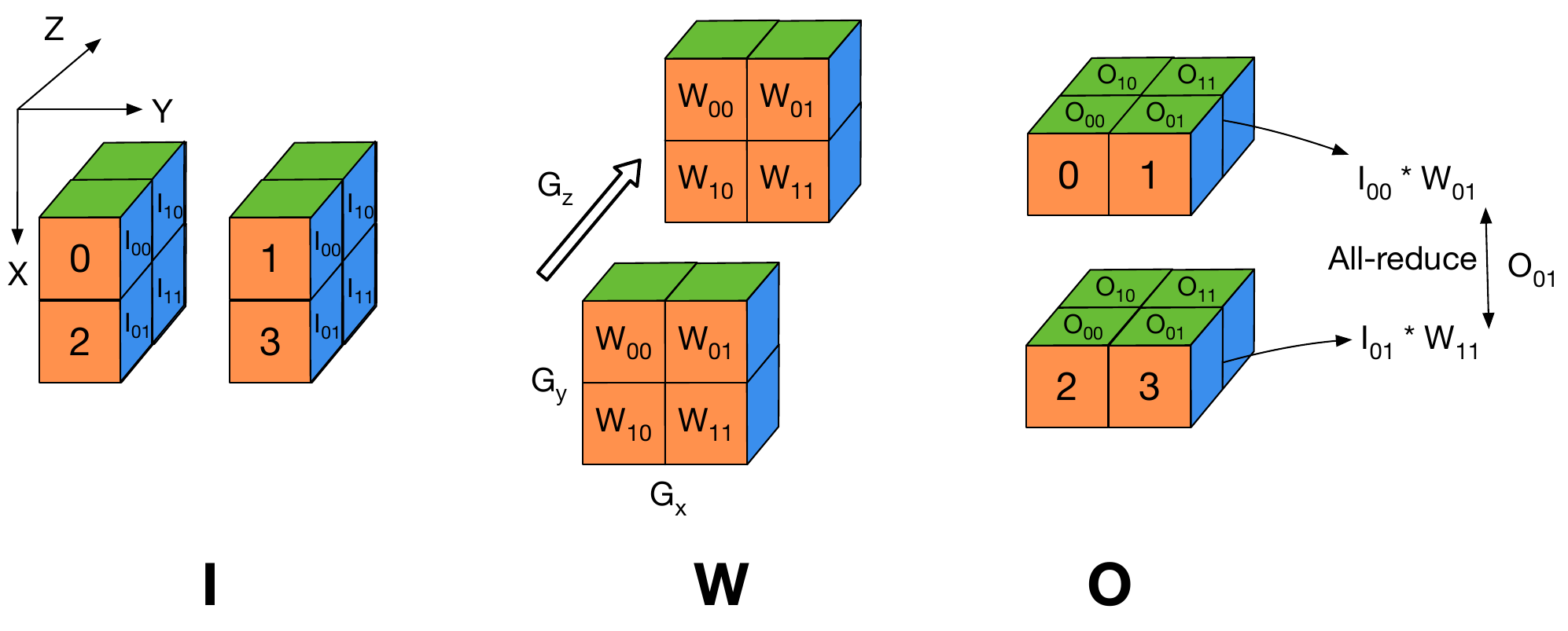}
      \caption{Parallelization of a matrix multiply in an FC layer with Agarwal's 3D parallel matrix
multiplication algorithm~\cite{agarwal-3d} on eight GPUs organized in a
$2\times2\times2$ topology. We use $G_x$, $G_y$, and $G_z$ to refer to
the number of GPUs along the three dimensions of the virtual grid topology.}
      \label{fig:schematic-agarwal-data-dist}
\end{figure}

We adapt Agarwal et al.'s 3D parallel matrix multiplication
algorithm~\cite{agarwal-3d}, for parallelizing our MMs. The 3D refers to
organizing the workers (GPUs) in a three-dimensional virtual grid.  So, we
organize the $G_{\mathrm{tensor}}$ GPUs further into a virtual 3D grid of
dimensions $G_x \times G_y \times G_z$
(Figure~\ref{fig:schematic-agarwal-data-dist}).  We do 2D decompositions of
both $I$ and $W$ into sub-blocks and map them to orthogonal planes of the 3D
grid. In the figure below, $I$ is distributed in the $XZ$ plane, and copied in the $Y$ dimension. $W$ is distributed in the $XY$ plane and copied along the $Z$ dimension. Once each GPU has a unique sub-block of I and W, it can compute a
portion of the $O$ matrix, which can be aggregated across GPUs in the $X$
direction using all-reduces.

We modify Agarwal's algorithm to reduce memory consumption, and instead of
making copies of $W$ along the $Z$-axis, we further shard $W$ along the
$Z$-axis and denote these sub-shards as $\hat{W}$.
Algorithm~\ref{alg:3d-tensor} presents the forward and backward pass operations
on GPU $g_{i,j,k}$, and we can observe that the sharding of $W$ results in
all-gather operations before the local matrix multiplication on each GPU can proceed.

\begin{algorithm}[h]
    {\small 
    \caption{Tensor parallel algorithm for ${g}_{i,j,k}$ in a $G_{x} \times
G_{y} \times G_{z}$ grid. Communication operations highlighted in blue.} 
    \label{alg:3d-tensor}
    \begin{algorithmic}[1]
    \setlength{\lineskip}{5pt}
    \Function{tensor\_parallel\_forward\_pass}{$I_{k,j}$, $\hat{W}_{j,i}$}
        \State  $W_{j,i}$ = $\Call{\comm{${\text{all-gather}}_{z}$}}{\hat{W}_{j,i}}$
        \State  $\hat{O}_{k,i} = I_{k,j} \times {W}_{j,i}$
        \State $O_{k,i}$ $\gets$ \Call{\comm{${\text{all-reduce}}_{y}$}}{$\hat{O}_{k,i}$}
        \State // Cache $I_{k,j}$ and $W_{j,i}$  for the backward pass
        \State \Return $O_{k,i}$
    \EndFunction
    \State
    \Function{tensor\_parallel\_backward\_pass}{$\frac{\partial L}{\partial O_{k,i}}$}
        \State Retrieve  $I_{k,j}$ and $W_{j,i}$  from cache 
        \State $\hat{\frac{\partial L}{\partial I_{k,j}}}$ $\gets$ $\frac{\partial L}{\partial O_{k,i}} \times {W}^{\top}_{j,i}  $
        \State $\frac{\partial L}{\partial I_{k,j}}$ $\gets$ \Call{\comm{${\text{all-reduce}}_{x}$}}{$\hat{\frac{\partial L}{\partial I_{k,j}}}$}
        \State ${\frac{\hat{\partial L}}{\partial \hat{W}_{j,i}}}$ $\gets$ $ I^{\top}_{k,j}  \times {\frac{\partial L}{\partial O_{k,i}}} $
        \State ${\frac{\partial L}{\partial \hat{W}_{j,i}}}$ $\gets$ \Call{\comm{$\text{reduce-scatter}_{z}$}}{$ \frac{\hat{\partial L}}{\partial \hat{W}_{j,i}} $}
        \State \Return $\frac{\partial L}{\partial I_{k,j}}$, ${\frac{\partial L}{\partial \hat{W}_{j,i}}}$
    \EndFunction
    \end{algorithmic}
    }
\end{algorithm}

In the forward pass, after the local (to each GPU) matrix-multiply on line 3,
we do an all-reduce to aggregate the output activations (line 4). In the
backward pass, there are two matrix multiplies on lines 11 and 13, and
corresponding all-reduce and reduce-scatter operations in lines 12 and 14 to
get the data to the right GPUs.

\vspace{0.08in}
\noindent\textbf{Parallelizing an entire network:} 
The approach of parallelizing a single layer in a deep neural network can be
applied to all the layers individually. Let us consider a 2-layer neural
network.  If we use Algorithm~\ref{alg:3d-tensor} to parallelize each layer,
the output $O$ of the first layer would be the input to the other. However,
notice in Figure~\ref{fig:schematic-agarwal-data-dist} that $O$ is distributed
across the 3D virtual grid differently than the input $I$. So to ensure
that the second layer can work with $O$, we would need to transpose its weight
matrix -- essentially dividing its rows across the $X$-axis and columns across
the $Y$-axis. This transpose needs to be done once at the beginning of
training.  Hence, to parallelize a multi-layer neural network, we simply
`transpose' the weights of every other layer by swapping the roles of the
$X$- and $Y$- tensor parallel groups.

Note that the 4D algorithm (data + 3D PMM) discussed in this section is a generalization of
various state-of-the-art parallel deep learning algorithms. For example, if one
were to employ only the $Z$ axis of our PMM algorithm to parallelize training, it
would reduce to Fully Sharded Data Parallelism (FSDP)~\cite{fsdp} and
ZeRO~\cite{sc2020zero}. Similarly, if we employ the $Z$ axis of 3D PMM and data
parallelism simultaneously, then our algorithm reduces to Hybrid Sharded Data
Parallelism~\cite{fsdp} and ZeRO++~\cite{wang2023zero}. If we use the $X$
axis of our 3D PMM algorithm along with the `transpose' scheme discussed in the
previous paragraph, our 4D algorithm reduces to Shoeybi et al.'s
Megatron-LM~\cite{megatronlm}. Finally, when all four dimensions of our 4D algorithm are being used, this is similar to a hybrid scheme that combines data parallelism, FSDP, and two-dimensional tensor parallelism.

\subsection{A Performance Model for Identifying Near-optimal Configurations}
When assigned a job partition of $G$ GPUs, we have to decide how to organize
these GPUs into a 4D virtual grid, and how many GPUs to use for data
parallelism versus the different dimensions of 3D parallel martix
multiplication. To automate the process of identifying the best performing
configurations, we have developed a performance model that predicts the
communication time of a configuration based on the neural network architecture,
training hyperparameters, and network bandwidths. Using these predictions, we
can create an ordered list of the best performing configurations as predicted by
the model. We describe the inner-workings of this model below.

We primarily focus on modeling the performance of the collective operations in
the code, namely all-reduces, reduce-scatters, and all-gathers.  We first list
the assumptions we make in our model:
\begin{itemize}
    \item \emph{Assumption-1:}  
    The ring algorithm~\cite{thakurimproving2003} is used for implementing the all-reduce, reduce-scatter, 
    and all-gather collectives.
    \item \emph{Assumption-2:} For collectives spanning more than one compute node, the ring is formed such that the number of messages crossing node 
    boundaries is minimized.
    \item \emph{Assumption-3:} The message sizes are large enough, and hence, message startup overheads can be 
    ignored. 
    In other words, if a process is sending a message of $n$ bytes, then we assumed that the transmission time is simply 
    $\frac{n}{\beta}$, where $\beta$ is the available bandwidth between the two processes.
    \item \emph{Assumption-4:} We only model the communication times and ignore the effects of any computation taking 
    place on the GPUs.  
    \item \emph{Assumption-5:} We assume the same peer-to-peer bidirectional bandwidth, $\beta_{\mathrm{inter}}$, between every 
    pair of nodes.
\end{itemize}

We use the analytical formulations in Thakur et al.~\cite{thakurimproving2003}
and Rabenseifner~\cite{rabenseifneroptimization2004} for modeling the
performance of ring algorithm based collectives.  Let $t_{\mathrm{AG},z}$
denote the time spent in the all-gather across the $Z$-tensor parallel groups
(line 2 of Algorithm~\ref{alg:3d-tensor}). Similarly, we use
$t_{\mathrm{RS},z}$, $t_{\mathrm{AR},y}$ and $t_{\mathrm{AR}, x}$ to refer to
the time spent in the collectives in lines 14, 4, and 12 respectively.
Similarly, we use $t_{\mathrm{AR}, \mathrm{data}}$ for the time spent in the
data parallel all-reduce. Then, we can model these times as follows,
\begin{align}
    t_{\mathrm{AG},z} &= \frac{1}{\beta}\times (G_{z}-1) \times \frac{k \times n}{G_{x} \times G_{y} \times G_{z}}  
    \label{eqn:layer-ag} \\ \nonumber \\
    t_{\mathrm{RS},z} &= \frac{1}{\beta}\times \left( \frac{G_{z}-1}{\mathrm{G_{z}}} \right) \times \frac{k \times n}{G_{x} \times G_{y}} 
    \label{eqn:layer-rs} \\ \nonumber \\
    t_{\mathrm{AR},y} &= \frac{2}{\beta} \times \left( \frac{G_{y}-1}{G_{y}} \right) \times \frac{m \times n}{G_{z} \times \mathrm{G}_{x}}
    \label{eqn:layer-ar-1} \\ \nonumber \\
    t_{\mathrm{AR}, x} &= \frac{2}{\beta} \times \left( \frac{G_{x}-1}{G_{x}} \right)
    \times \frac{m \times k}{G_{z} \times G_{y}}
    \label{eqn:layer-ar-2} \\ \nonumber \\
    t_{\mathrm{AR}, \mathrm{data}} &= \frac{2}{\beta} \times \left( \frac{G_{\mathrm{data}}-1}{G_{\mathrm{data}}} \right) 
    \times \frac{k \times n}{G_{x} \times G_{y} \times G_{z}}
    \label{eqn:layer-ar-3}  
\end{align}

The total communication time for a single layer, ${t_{\mathrm{comm}}}$ is simply
the sum of Equations~\ref{eqn:layer-ag} through~\ref{eqn:layer-ar-3}: 
\begin{align}
    t_{\mathrm{comm}} = t_{\mathrm{AG},z} + t_{\mathrm{RS},z} +  t_{\mathrm{AR},y} + t_{\mathrm{AR},x} 
    + t_{\mathrm{AR}, \mathrm{data}} \label{eqn:layer} 
\end{align}
For layers with `transposed' weight matrices as discussed at the end of
Section~\ref{sec:hybrid}, we need to swap the values of $G_x$ and $G_y$. 
And finally, to model the communication time for
the entire network, we apply Equation~\ref{eqn:layer} to all of its layers, and
take a sum of the times.

In the equations derived above, we made a simplifying assumption that all
collectives in our hybrid parallel method can achieve the highest peer-to-peer
bandwidth, denoted by ${\beta}$. However, since several collectives are often
in operation at once, the actual bandwidth achieved for a collective operation
among a group of GPUs depends on the placement of processes in our 4D virtual
grid to the underlying hardware topology (nodes and network)~\cite{solomonik:sc2011, bhatele:sc2012b, abdel-gawad:sc2014, bhatele:hipc2014}. For example,
process groups that are contained entirely within a node can experience higher
bandwidths than those containing GPUs on different nodes. Next, we
model the specific bandwidths used in Equations~\ref{eqn:layer-ag}
through~\ref{eqn:layer-ar-3}.

To model the process group bandwidths, we begin by assuming a hierarchical organization of process groups: 
$X$-tensor parallelism (innermost), followed by $Y$-tensor parallelism, $Z$-tensor parallelism, and
data parallelism (outermost). As a concrete example, if we have eight GPUs, and set $G_{x}=G_{y}=G_{z}=G_{\mathrm{data}}=2$, 
then the $X$-tensor parallel groups comprise of GPU pairs 
(0,1), (2,3), (4,5), and (6,7). Similarly, the $Y$-tensor parallel groups would comprise of GPU pairs (0,2),
(1,3), (4,6), and (5,7), and so on.
% for the $Z$ and $\mathrm{data}$ parallel groups.

Now let $\vec{G} = (G_{x}, G_{y}, G_{z}, G_{\mathrm{data}})$ be the tuple of our configurable performance 
parameters, arranged in order of the assumed hierarchy. Let 
$\vec{\beta} = (\beta_{x}, \beta_{y}, \beta_{z}, \beta_{\mathrm{data}})$ be the effective peer-to-peer bandwidths 
for collectives issued within these process groups. We use $\vec{\beta}_{i}$ and $\vec{G}_{i}$ to represent the $i^{\mathit{th}}$
elements of these tuples ($0 \leq i \leq 3$). Also, let $G_{\mathrm{node}}$ refer to the number of GPUs per node. Now let us 
model each $\beta_{i}$ i.e. the bandwidth available to the GPUs in the process groups at the $i^{\mathit{th}}$ level of the 
hierarchy.

\vspace{0.08in}
\noindent{\em Case 1: GPUs in the process group lie within a node} --
in our notation, this is the scenario when $\prod_{j=0}^{i}G_{j} \leq G_{\mathrm{node}}$. 

The bandwidth $\vec{\beta}_{i}$ is determined by two primary factors: (i) the
size of the $i$th process group, $G_{i}$, and (ii) the cumulative product of
the sizes of all preceding process groups, $\prod_{j=0}^{i-1}G_{j}$. Given that
the number of GPUs per node is typically small, the number of possible
scenarios is also small. Therefore, we can profile the bandwidths for all
potential configurations in advance and store this information in a database.
Specifically, we generate all possible two-dimensional hierarchies of process
groups $(G_{0}, G_{1})$ such that $G_{0} \times G_{1} \leq G_{\text{node}}$,
and then perform simultaneous collectives within the outer process groups of
size $G_{1}$ with a large message size of 1 GB. We record the achieved
bandwidths for this tuple in our database. Then, for a given model, when we
need the predicted bandwidths for the $i^{th}$ process group, we retrieve the
bandwidth recorded for the tuple $(G_{0} = \prod_{j=0}^{i-1}G_{j}, G_{1} =
G_{i})$.

\begin{figure*}[t]
    \centering
    \includegraphics[width=0.49\textwidth]{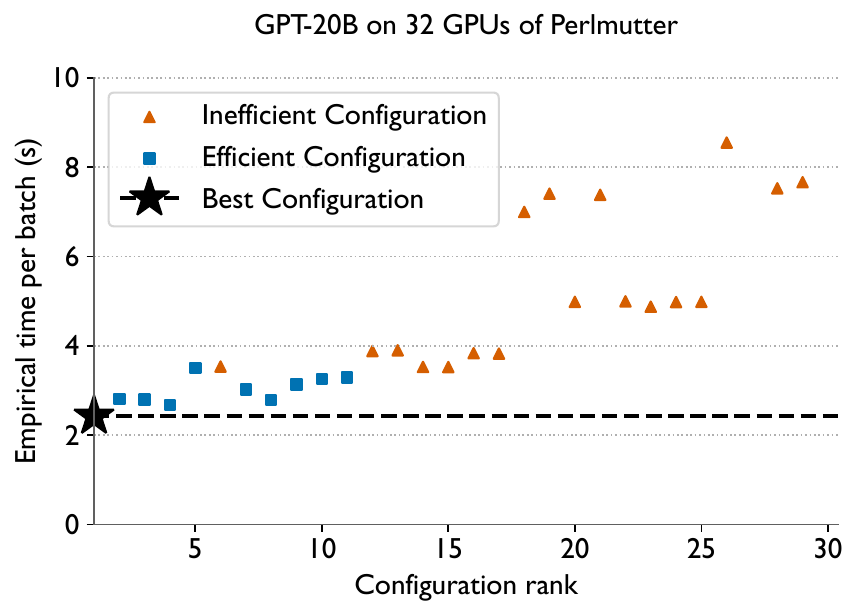}
    \includegraphics[width=0.49\textwidth]{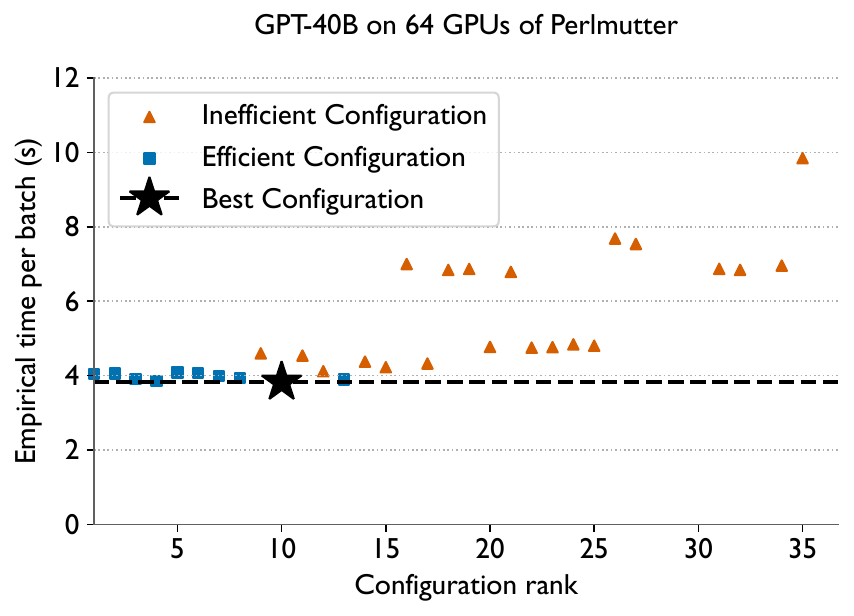}
    \caption{Plots validating the performance model by comparing the observed time per batch and the rank ordered by the model for two neural networks: GPT-20B (left) and GPT-40B (right).}
    \label{fig:rank-comm-model}
\end{figure*}

\vspace{0.08in}
\noindent{\em Case 2: GPUs in the process group are on different nodes} --
in our notation, this is the scenario when $\prod_{j=0}^{i}G_{j} > G_{\mathrm{node}}$. 

For process groups spanning node boundaries, the approach of recording all
possible configurations in a database is not feasible due to the vast number of
potential scenarios, given the large number of possible sizes of these groups
in a multi-GPU cluster.  Therefore, we develop a simple analytical model for
this scenario, which predicts the achieved bandwidths as a function of the
inter-node bandwidths ($\beta_{\mathrm{inter}}$), process group sizes
($\vec{G}$), and the number of GPUs per node ($G_{\text{node}}$).

\begin{figure}[h]
    \centering
       \includegraphics[width=0.9\columnwidth]{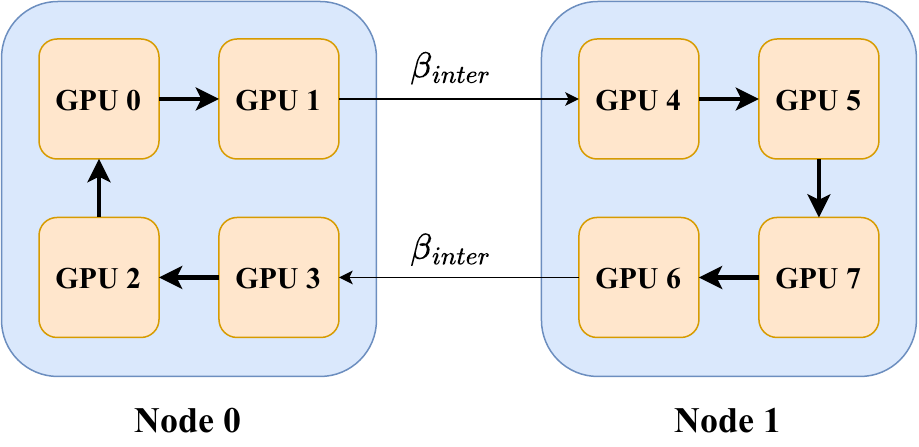}
       \caption{Creation of a ring among eight GPUs on two nodes for a collective communication operation (all-reduce/reduce-scatter/all-gather).}
       \label{fig:all-reduce-one}
\end{figure}

First, let's first explore two simple examples to build some intuition. In Figure~\ref{fig:all-reduce-one}, we demonstrate a 
scenario with a single process group spanning eight GPUs on two nodes, with four GPUs on each node. In this case, the ring 
messages crossing node boundaries (i.e. the link between GPUs 1 and 4, and between GPUs 6 and 3) will be the 
communication bottleneck. Since we assumed $\beta_{\mathrm{inter}}$ to be the bidirectional bandwidth between node pairs, we 
can set $\beta_{i}=\beta_{\mathrm{inter}}$.

\begin{figure}[h]
    \centering
       \includegraphics[width=0.9\columnwidth]{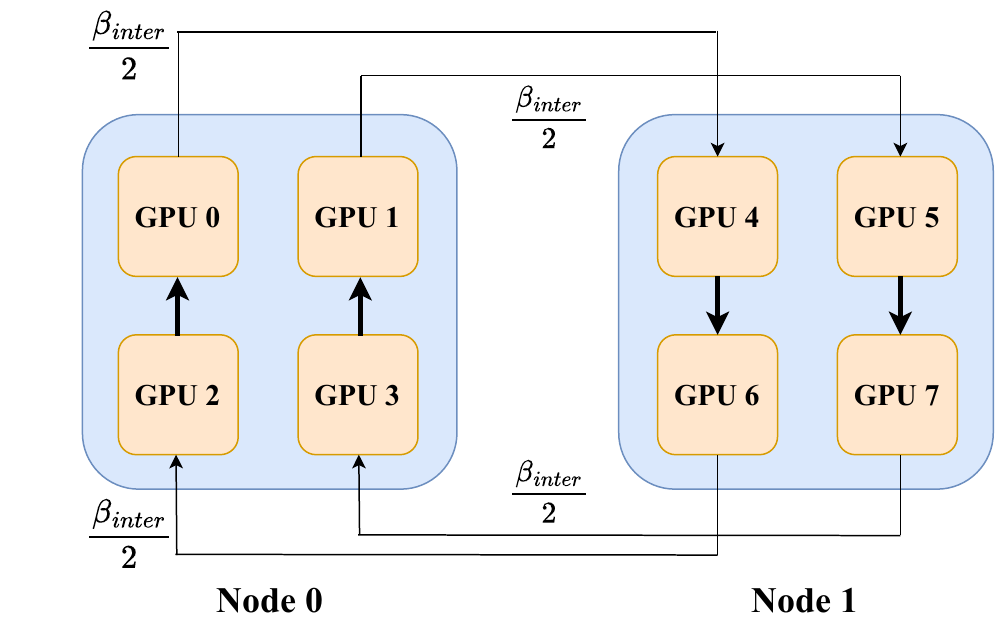}
       \caption{Two rings among four GPUs each across two nodes for performing collective operations simultaneously.}
       \label{fig:all-reduce-many}
\end{figure}

Another possible scenario is when there are multiple simultaneous collectives taking place between two nodes. For example, consider 
Figure~\ref{fig:all-reduce-many}, wherein GPUs $(0, 4, 6, 2)$ and GPUs $(1, 5, 7, 3)$ are executing two independent collectives using the ring algorithm
simultaneously. In this case, the available inter-node bandwidth will be shared between these two collectives and 
$\beta_{i}=\frac{\beta_{\mathrm{inter}}}{2}$. 

The first scenario occurs in the case when the process groups preceding the $i^{\mathit{th}}$ process 
group in the hierarchy are of size one, i.e. $G_{j}=1$  $\forall j < i$. Whereas the second scenario occurs in the case 
when at least one of these preceding process groups is of a size $>1$. In that case, we get multiple ring messages crossing node boundaries
and the bandwidth gets distributed between the rings. However, note that the maximum
reduction in the bandwidth is bounded by the total number of GPUs on each node, as there can't be more 
inter-node ring links than GPUs on a node. Equation~\ref{eqn:bandwidth} models all the scenarios to obtain the observed bandwidth:
\begin{equation}
\vec{\beta}_{i} = \
         \dfrac{\beta_{\text{inter}}}{\min \left( G_{\text{node}}, \prod_{j=0}^{i-1}G_{j} \right)} \label{eqn:bandwidth}
\end{equation}
We use this bandwidth term in Equations~\ref{eqn:layer-ag}
through~\ref{eqn:layer-ar-3} of our model. We use the model to create an
ordered list of configurations, and then we can pick the top few configurations
for actual experiments.

\vspace{0.08in}
\noindent{\bf Validating the Performance Model}: To validate the model, we
collect the batch times for all possible configurations of the 4D virtual grid
when training GPT-20B on 32 GPUs and GPT-40B on 64 GPUs of Perlmutter. Using
the observed batch times, we label the ten fastest configurations as `efficient'
and the rest as `inefficient'. When creating the validation plots, we rank the
configurations using the ordering provided by the performance model.
Figure~\ref{fig:rank-comm-model} shows the empirical batch times on the Y-axis
and the rank output by the performance model on the X-axis. The fastest
configurations should be in the lower left corner. We observe that nine out of
the top ten configurations predicted by the performance model are indeed
`efficient' as per their observed batch times. This shows that the model is
working very well in terms of identifying the fastest configurations.

\subsection{Automated Tuning of BLAS Kernels} \label{sec:blas-tune}
In deep neural networks, a significant portion of the computational workload is
matrix multiplications kernels or ``matmuls``, particularly in transformer
models. These matmuls can be performed in one of three main modes based on
whether the operands are transposed: NN, NT, and TN. Prior research has
highlighted that NT and TN kernels are often less optimized than NN kernels in
most BLAS libraries~\cite{shi2017tnvnn}. In our experiments, we found this
discrepancy to be more pronounced when running transformers with large hidden
sizes on the AMD MI250X GPUs of Frontier. For example, in the GPT-320B model
(described in Table~\ref{tab:setup-perf-gpt}), we observed that a matrix
multiply defaulting to the TN mode in PyTorch achieved only 6\% of the
theoretical peak performance, whereas other matmuls reached 55\% of the peak.

To address this issue, we implemented an automated tuning strategy in which,
during the first batch, each matmul operation in the model is executed in all
three modes (NN, NT, and TN) and timed. We then select the most efficient
configuration for each operation, which is subsequently used for the remaining
iterations. This tuning approach ensures that our deep learning framework,
AxoNN, avoids the pitfalls of using suboptimal matmuls that could significantly
degrade performance. For the aforementioned 320B model, our BLAS kernel tuning
approach successfully switches the poorly performing TN matmul with a nearly
8$\times$ faster NN matmul, thereby reducing the total time spent in
computation from 30.1 seconds to 13.19s! Note that for other models used in
Table~\ref{tab:setup-perf-gpt}, the speedups attained via tuning are relatively
modest (See Figure~\ref{fig:weak-scaling-breakdown}).

\subsection{Overlapping Asynchronous Collectives with Computation}

We use non-blocking collectives implemented in NCCL and RCCL on NVIDIA and AMD
platforms respectively. This enables us to aggressively overlap the collective
operations in AxoNN with computation, which can minimize communication
overheads.

\vspace{0.08in}
\noindent{\bf Overlapping All-reduces with Computation (OAR):}
In this performance optimization, we overlap the all-reduce across the
$X$-tensor parallel groups in the backward pass (Line 12 of
Algorithm~\ref{alg:3d-tensor}) with the computation in Line 13. Once this
computation has completed, we wait on the asynchronous all-reduce.
Note that for layers with `transposed' weight matrices, this communication happens across the $Y$-tensor parallel groups.  

\vspace{0.08in}
\noindent{\bf Overlapping Reduce-scatters with Computation (ORS):}
Next we overlap the reduce-scatters in the backward pass (line 14 of
algorithm~\ref{alg:3d-tensor}). The outputs of this reduce-scatter are the
gradients of the loss w.r.t.~the weights. These outputs are not needed until
the backward pass is completed on all the layers of the neural network and we
are ready to start issuing the all-reduces in the data parallel phase.
Exploiting this, we issue these reduce-scatters asynchronously and only wait on
them once all layers have finished their backward pass. This allows us to
overlap the reduce-scatter of one layer with the backward pass computations of
the layers before it.

\vspace{0.08in}
\noindent{\bf Overlapping All-gathers with Computation (OAG):}
Our next optimization overlaps the all-gather operations in the forward pass
(line 2 of Algorithm~\ref{alg:3d-tensor}) with computation. We observe that
this all-gather operation does not depend on any intermediate outputs of the
forward pass. Leveraging this, we preemptively enqueue the all-gather for the
next layer while the computation for the current layer is ongoing. At the start
of training, we generate a topological sort of the neural network computation
graph to determine the sequence for performing the all-gathers. Subsequently,
we execute them preemptively in this order.

\begin{figure}[h]
    \centering
      \includegraphics[width=\columnwidth]{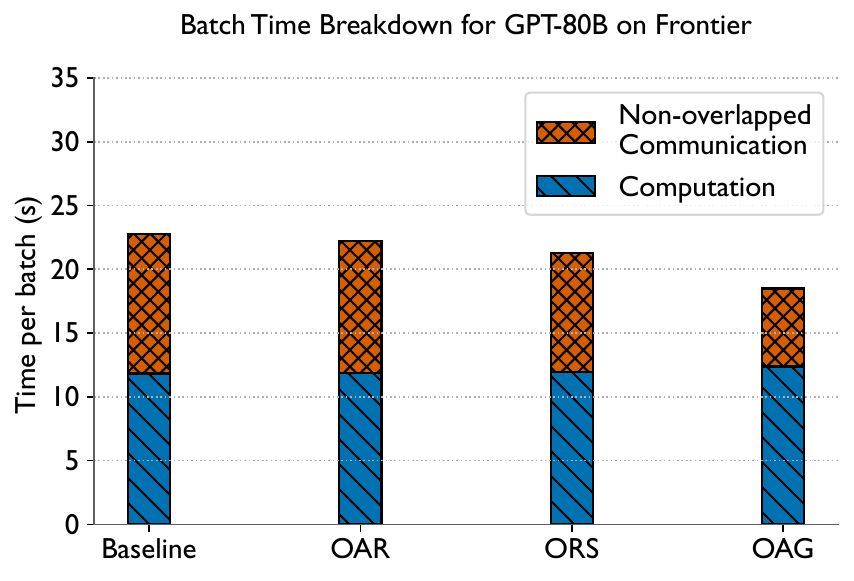}
      \caption{The impact of overlapping non-blocking collectives with
computation on the training times of different sized models on 8,192 GCDs of Frontier.}
\label{fig:opt}
\end{figure}

Figure~\ref{fig:opt} shows the performance improvements from the three
successive collective overlap optimizations (OAR: Overlap of all-reduces, ORS:
Overlap of reduce-scatters, and OAG: Overlap of all-gathers). The baseline here
refers to the scenario with no communication overlap. We also show the breakdown of the total time per
batch into computation and communication. As we can see, the times spent in
computation do not change significantly, however, the time spent in non-overlapped
communication reduces with successive optimizations, leading to an overall
speedup. For the 80B model in the figure, we see a performance improvement of 18.69\%
over the baseline on 8,192 GCDs of Frontier.

\section{How Performance Was Measured}
\label{sec:measurement}
% (Note that preference is given to performance actually measured [not
% projected], based on the entire application [including I/O] and with uniform
% precision.  Explain in detail if any portion of total runtime was not
% included in the measurements, if and where different precisions were used, or
% any attributes listed in Section 3 as “other”).

% what application(s) was used to measure performance (1 p max)

% system and environment where performance was measured (1 p max)

% \fix{2 pages max}

All of our innovations are implemented in an open-source framework called
AxoNN~\cite{singh:ipdps2022}, which can be integrated easily as a backend in
existing serial training codebases. This section provides details of the
experimental setup for benchmarking training performance using AxoNN.

\subsection{Applications: Model Architecture Details} \label{sec:setup-desc}
  
We evaluate the effectiveness of our implementation by conducting experiments
on a well-known neural network architecture: Generative Pre-trained Transformer
(GPT)~\cite{gpt-3}. The GPT architecture is a popular transformer
architecture~\cite{transformer} that has been used to train several large
language models~\cite{megatron-turing-nlg-530b, bloom176b, gpt-3, gpt-2}.
Table~\ref{tab:setup-perf-gpt} presents the sizes of the different model
architectures used in the experiments, and their important hyperparameters. Due
to the extremely large activation memory requirements of training GPT models,
we turn on activation checkpointing~\cite{chen2016training}. Additionally, we
employ mixed precision (bf16/fp32) for all our training runs. We use bf16 since
it has been shown to achieve the same performance and stability as
fp32~\cite{bfloat16studyfordl}, and it maintains the same range as fp32. This
makes it a suitable choice over fp16, which has  been known to be numerically
unstable for LLM training.

\begin{table}[h]
  \centering
  \caption{\label{tab:setup-perf-gpt} Architectural details of the GPT-style transformers~\cite{gpt-3} used in the performance experiments.
  }
  \begin{tabular}{lrcrr}
  \toprule
  Model      & \# Parameters & \# Layers  & Hidden-Size &\# Heads  \\ \midrule
  GPT-5B     &   5B & 24      & 4096        & 32   \\
  GPT-10B    &  10B & 32      & 5120        & 40   \\
  GPT-20B    &  20B & 32      & 7168        & 56   \\ 
  GPT-40B    &  40B & 38      & 9216        & 72    \\
  GPT-60B    &  60B & 56      & 9216        & 72    \\
  GPT-80B    &  80B & 42      & 12288       & 96    \\
  GPT-160B   & 160B & 84      & 12288       & 96    \\ 
  GPT-320B   & 320B & 96      & 16384       & 128    \\
  GPT-640B   & 640B & 192      & 16384       & 128    \\ \bottomrule
  \end{tabular}
\end{table}

On Perlmutter, we use the sequential model training code from the Megatron-LM
codebase~\cite{megatronlm-2}, and parallelize it using AxoNN. However, on
Frontier, we observed training instabilities with Megatron-LM, and switched to
using LitGPT~\cite{litgpt-2023} for the model architectures on Frontier and
Alps. We parallelized LitGPT also using our 4D implementation in AxoNN.  We
conduct weak scaling experiments with the GPT-3 models, ranging from 5 billion
to 320 billion parameters.  We also conduct strong scaling experiments on
Frontier using the 80 billion and 640 billion parameter models to predict the
time-to-solution for 2 trillion tokens.

\subsection{Systems and Environments} \label{sec:sys-and-env}

Our experiments were conducted on three supercomputers, Perlmutter at
NERSC/LBL, Frontier at OLCF/ORNL, and Alps at CSCS. Each node on Perlmutter is
equipped with four NVIDIA A100 GPUs, each with a DRAM capacity of 40 GB. On
Frontier, each node has four AMD Instinct MI250X GPUs each with a DRAM capacity
of 128 GB. Each MI250X GPU is partitioned into two Graphic Compute Dies (GCDs)
and each 64 GB GCD can be managed independently by a process. On Alps, each
node has four GH200 Superchips, where each H100 GPU has a DRAM capacity of 96
GB. Nodes on all systems have four HPE Slingshot 11 NICs, with each NIC capable
of bidirectional link speeds of 25 GB/s. 

In our Perlmutter experiments, we use CUDA 11.7, NCCL 2.15.5, and PyTorch 1.13. 
% We tried a newer build of PyTorch 2.1.0 with CUDA 12.0 and NCCL 2.18 but that
% led to a significant performance degradation of 25\% across our runs.
On Frontier, we use PyTorch 2.2.1 with ROCm 5.7 and RCCL 2.18.6.  On Alps, we
use PyTorch 2.4.0 with CUDA 12.5.1 and NCCL 2.22.3. On all the systems, we use
the AWS OFI plugin (NCCL or RCCL) which enables us to use libfabric as the
network provider on the Slingshot network, and provides high inter-node
bandwidth.  We want to note here that several runs on Perlmutter and Alps were
done in a system-wide reservation, and even so, we noticed significant
run-to-run performance variability. This was most likely due to network
congestion~\cite{bhatia:cgf2018} or file-system
degradation~\cite{mubarak:cluster2017} impacting performance.

\subsection{Evaluation Metrics}

In all our experiments, we run the training loop for ten iterations (batches),
and report the average time per iteration (batch) for the last eight iterations
to account for any performance variability due to initial warmup. We calculate
half precision flop/s (often called ``model flops'') using Narayanan et al.'s
analytical formulation~\cite{megatronlm-2} for the number of floating point
operations in a transformer model. We did a small experiment to verify that
this formulation matches the total number of floating point operations measured
by Nsight Compute, an empirical tool. We compare this number against the
theoretical (vendor advertised) peak performance of each GPU (312 Tflop/s per
GPU on Perlmutter, 191.5 Tflop/s per GCD on Frontier, 989 Tflop/s per GPU on
Alps), and report the achieved percentage of peak as well as the total
sustained bf16 flop/s.

Since the vendor advertised peak performance is often not practically
achievable, we also ran a simple GEMM benchmark on 1 GPU/GCD of
Perlmutter/Frontier to gather empirically observed peak flop/s.  We invoked
equivalent cuBLAS and rocBLAS kernel calls to multiply two bf16 square matrices
with dimensions ranging from 1024 to 65536.
% For each matrix dimension, we calculate the total number of floating point
% operations using the formula 2MKN. To measure the time taken, we repeat the
% multiplications for each dimension 100 times and take the average time of the
% last 75. 
On Perlmutter, the highest sustained flop/s for matrices of dimensions of 32768
$\times$ 32768 is 280 Tflop/s (90\% of peak). On Frontier, the highest
sustained flop/s is 125 Tflop/s on 1 GCD (65\% of peak) for the same matrix
dimensions. For Alps, we referred to a GH200 benchmark guide from NVIDIA that
reported a sustained performance of 813 Tflop/s (82\% of peak).  These numbers
show that the vendor advertised peak performance is almost always not
achievable in practice.  In our evaluation, we also report the \% of peak
empirical performance achieved by our implementation using the numbers
mentioned above.

\section{Performance Results}
\label{sec:results}
% include scalability (weak and strong), time to solution, efficiency (of
% bottleneck resources), and peak performance

% \fix{2 pages max}

We now discuss the results of our performance benchmarking
experiments described in Section~\ref{sec:measurement}.

\subsection{Weak Scaling Performance}

\begin{figure}[t]
  \centering
    \includegraphics[width=\columnwidth]{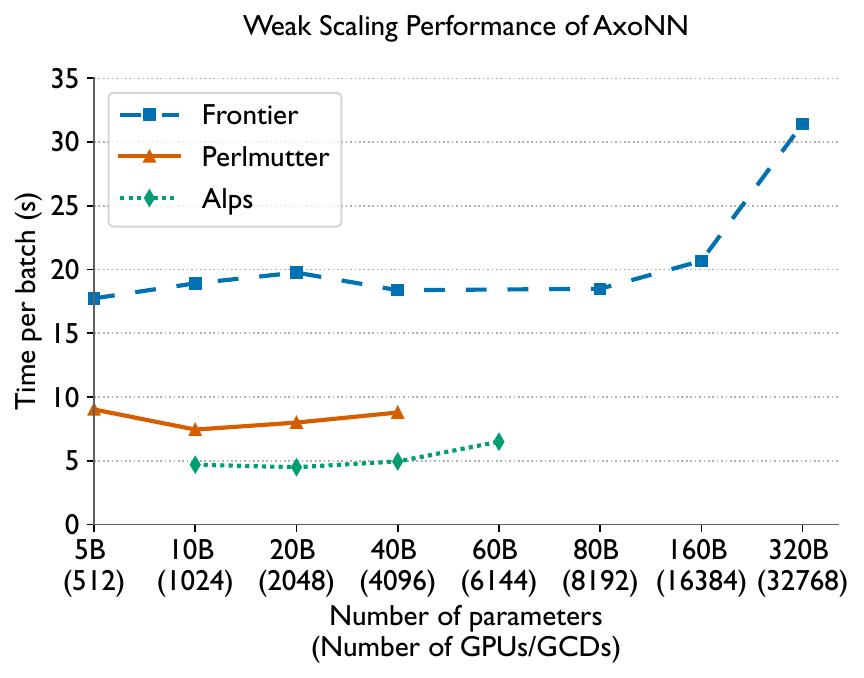}
    \caption{Weak scaling performance (time per batch or iteration) of AxoNN on Frontier, Perlmutter,
    and Alps for models with 5 to 320 billion parameters.}
    \label{fig:weak-scaling-time}
\label{fig:weak}
\end{figure}

We first present the weak scaling performance of AxoNN on Perlmutter, Frontier
and Alps using GPT-style transformers as the application in
Figure~\ref{fig:weak-scaling-time}.  We observe that on all three systems,
AxoNN achieves near-ideal weak scaling up to 4096 GPUs/GCDs. This is
particularly promising because most large-scale LLM training falls within this
hardware range. When running the 60B model on 6144 H100 GPUs of Alps, we see a
small reduction in efficiency -- 76.5\% compared to the performance on 1024
GPUs.

Since Frontier has a significantly large number of GPUs than the other two
platforms, we scaled AxoNN on Frontier to 32,768 GCDs. We see near perfect weak
scaling up to 8,192 GCDs with a significantly high efficiency of 88.3\%
(compared to the performance on 512 GCDs). Although our weak performance drops
at 16,384 GCDs, we are still able to sustain an efficiency of 79.02\%.
However, with rising overheads of communication, there is a notable decline in
our performance on 32,768 GCDs, and a corresponding drop in efficiency to
53.5\%.

\begin{figure}[h]
  \centering
    \includegraphics[width=\columnwidth]{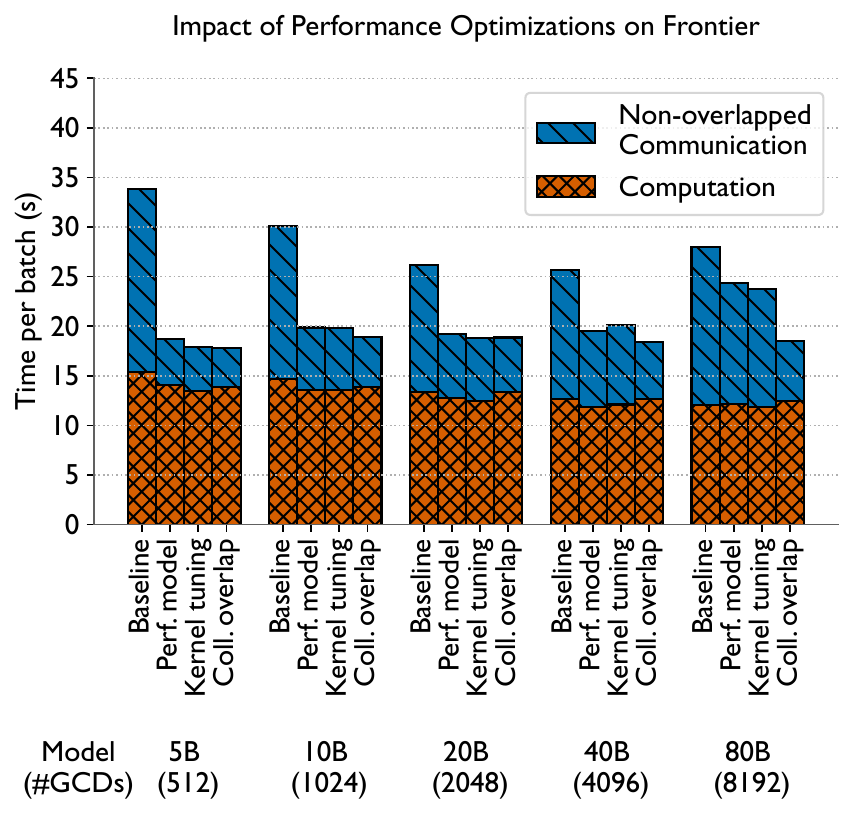}
    % \includegraphics[width=\columnwidth]{figs/weak_scaling_flops.pdf}
    % \caption{Time (left) total flop/s (right)}
    \caption{The impact of our performance optimizations on weak scaling of GPT
models.  For the bars labeled ``Perf model'', we use the best out of the top-10
configurations suggested by our communication model. For the bars labeled
``Kernel Tuning'' and ``Comm Overlap'', we enable our matrix multiplication
tuning and communication overlap optimizations.}
    \label{fig:weak-scaling-breakdown}
\label{fig:weak}
\end{figure}

We used timers to gather breakdowns of the time per batch into computation and
non-overlapped communication to better understand the impact of the performance
optimizations described in Section~\ref{sec:innovations}.  We present these
results in Figure~\ref{fig:weak-scaling-breakdown}, for some model sizes
running on 512--8,192 GCDs of Frontier.  As a baseline, we use a configuration
of AxoNN that corresponds to a hybrid of 1D tensor parallelism within node
(similar to Megatron-LM~\cite{megatronlm}) and hybrid sharded data parallelism
across nodes (similar to FSDP~\cite{fsdp,wang2023zero}).

We observe that using the 3D parallel matrix multiplication and performance
model to select the best configuration results in significant performance
improvements of 13-45\% over the baseline.  Most of the improvement comes from
a significant reduction in communication times. For the models in the plot, the
improvements in the batch times due to our BLAS kernel tuning are relatively
modest (2--4\%). Finally, the improvement from our overlap optimizations is
largest for the largest model in this series i.e.  80B on 8192 GCDs. In this
case, we observe a 22\% reduction in the batch times!  This is expected because
the overheads of communication tend to increase with scale and subsequently the
benefits of our overlap optimizations become more pronounced.

\begin{figure}[h]
  \centering
    \includegraphics[width=\columnwidth]{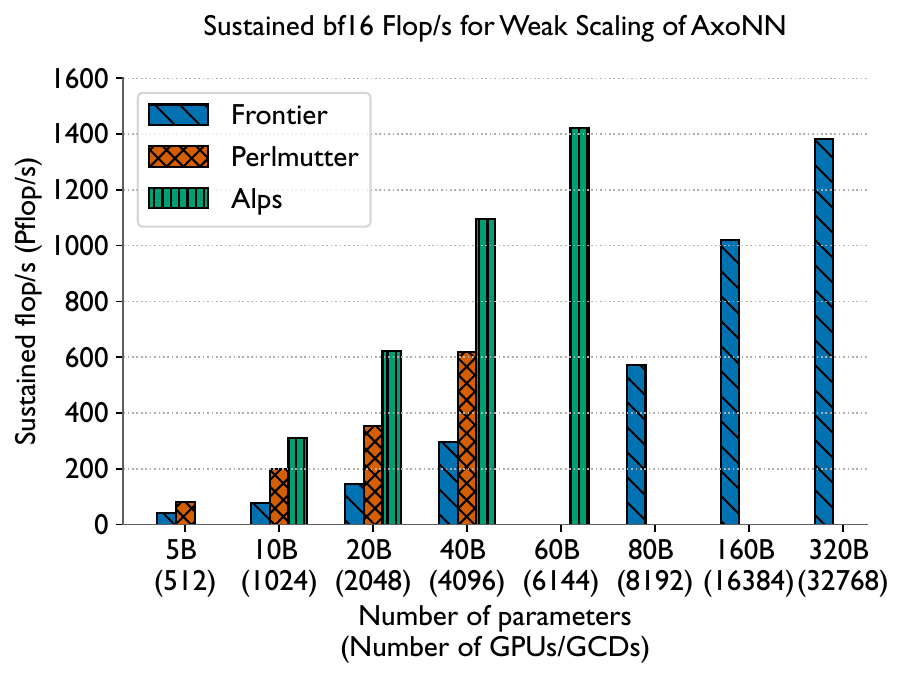}
    \caption{Sustained flop/s on different platforms. The FLOP count
is calculated analytically for all the matrix multiplication kernels in the
code.}
\label{fig:flops}
\end{figure}

\subsection{Sustained floating point operations per second (flop/s)}

Next, we examine the floating-point operations per second (flop/s) achieved by
AxoNN. In Figure~\ref{fig:flops}, we present the total bf16 flop/s sustained by
AxoNN in our weak scaling experiments on Perlmutter, Frontier and Alps. In
Table~\ref{tab:flops}, we also show our sustained flop/s as a percentage of the
vendor advertised and empirical obtained peak flop/s. As discussed in
Section~\ref{sec:sys-and-env}, we use 280 Tflop/s, 125 Tflop/s, and 813 Tflop/s
as the empirical peak bf16 flop/s for an A100 GPU, an MI250X GCD and an H100
GPU respectively.

On Perlmutter, we observe that AxoNN consistently sustains 
50\% or higher fraction of the advertised peak of 312 Tflop/s per GPU.
% In fact, on 1024 GPUs
% we see a significant increase to 60\% of the peak!
As a result of our near
perfect weak scaling, we observe that the sustained flop/s also increase
linearly from 80.8 Pflop/s on 512 GPUs by nearly $8
\times$ to 620.1 Pflop/s on 4096 GPUs. Since
the advertised and empirical peak bf16 flop/s of an A100 GPU are close
(312 vs.~280 Tflop/s), our \% flop/s numbers 
are also in the same ball park. 

\begin{table}[h]
  \centering
  \caption{Sustained flop/s for weak scaling on Perlmutter, Frontier and Alps.~\label{tab:flops}}
  \begin{tabular}{crrrcc}
  \toprule
  & \multicolumn{1}{c}{\begin{tabular}[c]{@{}c@{}}\# GPUs\\/ GCDs\end{tabular}} & Model & \multicolumn{1}{c}{\begin{tabular}[c]{@{}c@{}}Total \\ Pflop/s\end{tabular}} & \multicolumn{1}{c}{\begin{tabular}[c]{@{}c@{}}\% of \\ Advertised Peak\end{tabular}} & \multicolumn{1}{c}{\begin{tabular}[c]{@{}c@{}}\% of \\ Empirical Peak\end{tabular}} \\ \midrule
  \multirow{4}{*}{\begin{tabular}{@{}c@{}}\rotatebox[origin=c]{90}{Perlmutter}\end{tabular}} & 512 & 5B & 80.8 & 50.6 & 56.2  \\
  & 1024 & 10B & 197.8 & 61.9 & 68.8 \\
  & 2048 & 20B & 352.5 & 55.2 & 61.3 \\
  & 4096 & 40B & 620.1 & 48.5 & 53.9 \\ \midrule
  \multirow{7}{*}{\begin{tabular}{@{}c@{}}\rotatebox[origin=c]{90}{Frontier}\end{tabular}} & 512   & 5B   & 40.4  & 41.1 & 63.3 \\
  & 1024  & 10B  & 77.3  & 39.3 & 60.4 \\
  & 2048  & 20B  & 145.7 & 37.0 & 57.0 \\
  & 4096  & 40B  & 295.9 & 37.6 & 57.9 \\
  & 8192  & 80B  & 571.4 & 36.3 & 56.0 \\ 
  & 16384 & 160B & 1019.9 & 32.4 & 49.9 \\ 
  & 32768 & 320B & 1381.0 & 22.0 & 33.8 \\ \midrule
  \multirow{4}{*}{\begin{tabular}{@{}c@{}}\rotatebox[origin=c]{90}{Alps}\end{tabular}} & 1024   & 10B   & 310.0  & 30.6 & 37.3 \\
  & 2048  & 20B  & 621.6  & 30.7 & 37.4 \\
  & 4096  & 40B  & 1095.8 & 27.0 & 33.0 \\
  & 6144  & 60B  & 1423.1 & 23.4 & 28.6 \\ \bottomrule
  \end{tabular}
\end{table}

% AxoNN achieves near-perfect weak scaling in terms of sustained FLOP/s on Frontier between 512 and 4096 GCDs. This translates to a 
% throughput of around 40.5\% of the advertised peak performance.
On Frontier, in the 512 to 4,096 GCD range, AxoNN achieves near-perfect weak
scaling in terms of sustained flop/s which translates to a throughput of around
40\% of the advertised peak performance. Notably, this is a significant
improvement over Yin et al.~\cite{yin2023forge} and Dash et
al.~\cite{dash2023optimizing} -- they achieved a peak of only 30\% in a similar
range of GCDs, model sizes, and batch sizes on Frontier.  AxoNN continues to
scale well up to 8,192 GCDs, sustaining 36.3\% of the peak and 571.4 Pflop/s in
total. Beyond this scale, we start observing scaling inefficiencies. On 16,384
GCDs, we achieve 32.4\% of the peak, which amounts to 1.02 Exaflop/s in total.
Finally on 32,768 GCDs, our performance drops to 22\% of the peak and a total
flop/s of 1.381 Exaflop/s. In Section~\ref{sec:measurement}, we mentioned
a significant difference between the advertised peak and the empirically
measured peak on a single MI250X GCD (192 vs.~125 Tflop/s). As a result, there
is a large difference between AxoNN's flop/s expressed as a percentage of the
advertised peak versus the empirical peak. For instance on 32,768 GCDs, these
numbers are 22.0\% and 33.8\% respectively. 

On Alps, we observe a similar trend as Perlmutter, with AxoNN consistently
sustaining \tweakedsim30\% of the advertised peak up to 4096 GPUs. At 6144
GPUs, we see a slight drop to 23.42\% of peak.  At 6144 GPUs, we achieve our
highest sustained flop/s of 1423.10 Pflop/s across all three machines.

\subsection{Predicted Time-to-solution}

\begin{figure}[h]
  \centering
    \includegraphics[width=\columnwidth]{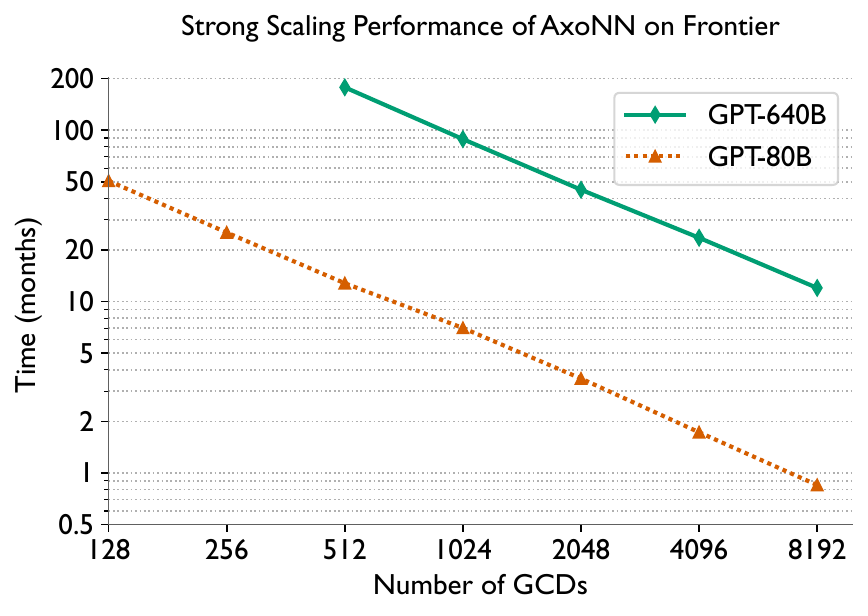}
    \caption{Strong scaling showing expected time-to-solution on Frontier. Using the average time per iteration, we predict the 
    training times for GPT-80B and GPT-640B on 2T tokens for various GCD counts.}
    \label{fig:strong}
\end{figure}

The training of state-of-the-art large language models (LLMs) presents a
significant computational challenge due to two key factors. First, the models
themselves are large, with current state-of-the-art models comprising hundreds
of billions of trainable parameters. Second, LLMs are trained on massive and
continually expanding text corpora, often containing trillions of tokens. In
this section, we show how AxoNN can significantly reduce the time-to-solution
of training such state-of-the-art LLMs on large text corpora. To demonstrate
this, we pick the 80B and 640B parameter GPT models from
Table~\ref{tab:setup-perf-gpt} and collect the per iteration times at various
GCD counts.  We run the 80B model on 128 to 8,192 GCDs on Frontier, and the
640B model on 512 to 8,192 GCDs. We then extrapolate the batch times to
estimate the time it would take to rain these models to completion i.e.~to
ingest two trillion tokens.  These time-to-solution results are presented in
Figure~\ref{fig:strong}. Note that both the model size and the number of tokens
are representative of modern LLM training setups such as Meta's
Llama~\cite{touvron2023llama}. 

As the plot shows, training an 80B model on 128 GCDs will take 50 months or
more than four years. This emphasizes the critical role of large-scale
parallelism in LLM training.  As we scale to more GCDs, we see the expected
time to solution drop almost linearly till 8,192 GCDs. Our estimate for the
total training time of the 80B model on 8,192 GCDs is a much more reasonable
25.5 days.  For the 640B model, even a much larger GCD count of 512 GCDs is
impractical, with the estimated time-to-solution amounting to 14 years.
However, on 8,192 GCDs, the estimated total training time is 15 months, which
is an 11$\times$ improvement.  For both models, this amounts to a strong
scaling efficiency of more than 90\%. These experiments underscore AxoNN's
efficacy in significantly reducing the pre-training time for cutting-edge LLMs
trained using massive datasets. By enabling faster training cycles on large
scale multi-GPU clusters such as Frontier and Alps, AxoNN has the potential to
accelerate the overall pace of LLM research and development.

\section{Implications}
\label{sec:implications}
% implications for future systems and applications

A scalable training framework such as AxoNN and access to large supercomputers
such as Frontier and Alps can enable studying properties of LLMs at scales that
were impossible before. Below, we present a study on the behavior of
memorization by large language models.

\subsection{Memorization of Training Data by Large Language Models}

A growing body of work has shown that language models memorize a portion of
their training data and can reproduce this training data at inference time
\cite{carlini2023quantifying}. The ability of LLMs to reproduce training data
has become a flashpoint for the AI community, as it poses major privacy and
legal risks for commercial models \cite{grynbaum2023times,
carlini2021extracting,carlini2023quantifying}.
% This effect grows rapidly as a function of the number of times a sequence is
% repeated in the data but findings also suggest that, memorization rates also
% climb with parameter count.

It is thought that memorization is largely due to training data repetition, and
it may be mitigated by dataset deduplication. Other factors such as data
structure and model size may play a factor, but the issue is not well
understood because public experiments have been constrained to smaller models
(e.g.~the popular Llama-2 7 billion parameter model~\cite{touvron2023llama})
with limited capacity and correspondingly small rates of memorization
\cite{carlini2023quantifying,biderman2023pythia}. As we observe below, the
ability to memorize entire documents emerges only for large model
sizes. Further, we hypothesize that models above a certain size threshold
may exhibit \textit{catastrophic memorization}, in which documents are
memorized immediately in one single pass. When training a model above this size
limit, even perfectly deduplicated datasets may still result in privacy and
copyright leaks.

By creating scalable, user-friendly and portable access to model parallelism,
AxoNN unlocks the potential for training and fine-tuning much larger models
under commodity computing constraints using sequential LLM training codebases.
This creates a scientific laboratory where large-model phenomena
such as memorization can be publicly reproduced and studied. It also raises the
ability of many practitioners to fine-tune large models on domain-specific
data, expanding the need to understand memorization risks.

\subsection{Experimental Setup: Training Llama models on Wikipedia}

We design a targeted set of continued pre-training experiments to quantify the
relationship between model size and memorization. We consider the Llama family
of LLMs with publicly available pre-trained weights, and use the AxoNN infused
LitGPT framework (introduced in Section~\ref{sec:setup-desc}) to parallelize
the models. Our experiments start with pre-trained checkpoints for the
TinyLlama-1B model~\cite{zhang2024tinyllama}, the 7B, 13B, and 70B parameter
models in the Llama 2 family~\cite{touvron2023llama} and the 8B, 70B, and 405B
parameter models from the recent Llama 3.1 release~\cite{dubey2024llama}. We
train on English text data from Wikipedia with varying levels of repetition to quantify how
memorization depends on model scale.

We train on English Wikipedia pages with $2048$ tokens or more. The articles
are randomly placed into one of four disjoint ``buckets,'' each with 200 articles. During
training, the first three buckets are repeated for 1, 4, or 6 ``epochs'' (one
pass over every page in the bucket) respectively. The fourth bucket is a
control group to measure baseline preexisting memorization from pre-training,
and we do not perform any further training on the pages in the fourth bucket. After training is
complete, we prompt the model with the beginning of each training sequence, and
let the model write the last 50 tokens. We consider a sequence memorized if the
model perfectly reproduces the correct 50 tokens.

%For each model size, we train on three buckets of Wikipedia pages for $1$, $4$
%and $6$ epochs respectively\footnote{The actual epoch counts are $1$, $\sim4.4$
%	and $\sim6.6$ epochs due to the stochastic nature of our batch sampling code.}.
We train the $1$B, $7$B, and $8$B models on eight GCDs of Frontier using
$8$-way $Z$-tensor parallelism (i.e. $G_{z}=8$), the $13$B model using $16$
GCDs, the $70$B models using $64$ GCDs, and the $405$B model using $128$ GCDs,
each with a corresponding level of $Z$-tensor parallelism. The total batch size
is fixed at $128$ samples for all model sizes. In the case of smaller models,
lower level of tensor parallelism is needed, so data parallelism is used to
utilize the remaining GPUs.  We warm up each model for $50$ steps, increasing
the learning rate to $3\times10^{-4}$ on the non-bucketed Wikipedia pages, and
then inject the three buckets of target data over the next $50$ steps of
training while decaying the learning rate to $3\times10^{-5}$. We report
memorization for each bucket separately, and also for the held-out (``$0$ Ep'')
control bucket.

% trim = left, bottom, right, top
\begin{figure*}[t]
	\centering
	\includegraphics[width=0.49\textwidth]{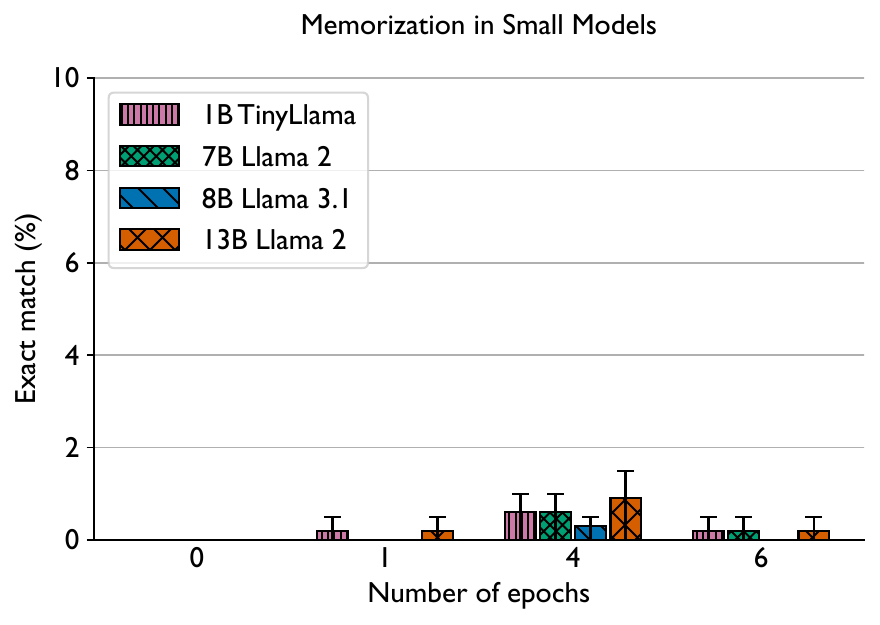}
	\includegraphics[width=0.49\textwidth]{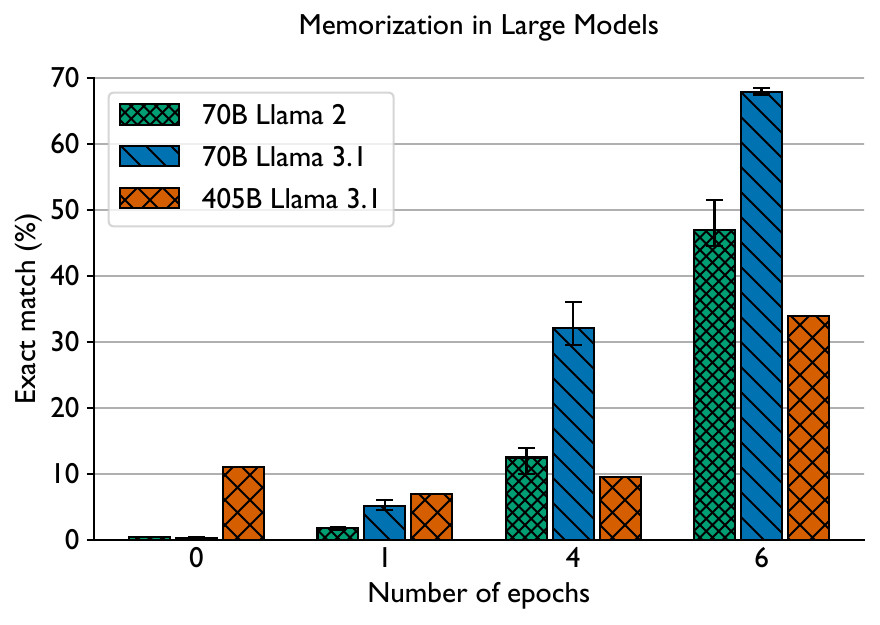}
    \caption{Memorization as a function of parameter count and epochs
(repetitions of the training data). For each model size, we show the ``Exact
Match'' rate at which the model correctly reproduces the last $50$ tokens of
articles after being trained on them for various numbers of epochs.
\textbf{(Left)}  Memorization is difficult to observe for small models.
\textbf{(Right)} The ability to efficiently memorize emerges at larger models
scales.  We see that a 70B model is even capable of {\em catastrophic
memorization}, as it memorized entire documents after seeing them just once.
For models with parameter counts in the
$1$B-$13$B range, we report the average over five trials, for $70$B, we report
the average over three trials, and for $405$B we report a single trial. Error
bars depict the min and max observed scores.}
    \label{fig:mem-results}
\end{figure*}

\subsection{Results: Catastrophic Memorization as a Function of Model Size}

% Chart data is here if you want to cite more specific numbers:
% figs/data_mem_raw_update/gordon_bell_update_2024-08-08_17-59-04.csv
% non_member=0Ep,bucket3=1Ep,bucket4=4Ep,bucket5=6Ep

Figure~\ref{fig:mem-results} shows the impact of parameter count and number of
epochs on exact memorization under otherwise identical conditions. At the
$1$B-$13$B scale (left plot), training for up to six epochs causes memorization of less
than $1\%$ of the $200$ documents on average. However, we observe that the $70$B
models and the 405B model are capable of significant memorization (right plot). After just six passes over the
data, the $70$B Llama 2 and $70$B Llama 3.1 models memorize $\mathbf{47\%}$ and
$\mathbf{67\%}$ of documents on average respectively. Furthermore, we observe
catastrophic memorization behavior starting at the $70$B scale; roughly 5\% of
documents are memorized in just one single pass.

Moving to the $405$B scale, we make several surprising observations. This model
had already memorized over 10\% of the control documents (see the bars labeled
``$0$ Ep'') before our experiment even began, showing that the ability to
memorize and retain documents during pre-training has emerged at this scale.
While Wikipedia pages were certainly included in the training corpus of the
Llama 3.1 series of models, only this largest model in the family exhibits such
non-trivial levels of memorization without further continued training.
Counterintuitively, we note that the rate of memorization of the $405$B model
during continued training was slower than that of the $70$B model. This is
likely because we used one set of hyperparameters for all models, and extreme
scales likely require different hyperparameters for optimal learning.

% \fix{405B caveats}
% @tomg other caveats not described above

% memorization rate appears to go down after we train for 1 and 4 epochs versus the number at 0.
% because we dont track the "diff" we cant tell the difference between the 405 forgetting some sequences versus learning some new ones.
% We couldn't run the model in the same precision setting. It is bf16-true, all other models are bf16-mixed. could matter more for very large deep models with loss of gradient precision when trying to fit tokens perfectly in a few passes.

\subsection{Results: Goldfish Loss Stops Memorization in its Tracks}

% Tom will rewrite this in his way I am sure :]
Observing extreme levels of memorization for models at the $70$B parameter
scale and above, we deploy a recently proposed technique for mitigating
memorization in large language models. Language model training
minimizes the expected cross-entropy between the language model's next-token
distribution and the true tokens as they appear in the training corpus. The
\textit{Goldfish Loss}~\cite{hans2024goldfish} technique introduces a mask such that some
tokens in any given training sequence are randomly omitted from the loss
computation. The model cannot memorize the masked tokens, and must ``guess''
them when trying to reproduce a training sequence at inference time, making it
very unlikely that long sequences can be exactly reproduced.

\begin{figure}[h]
	\centering
	\includegraphics[width=\columnwidth]{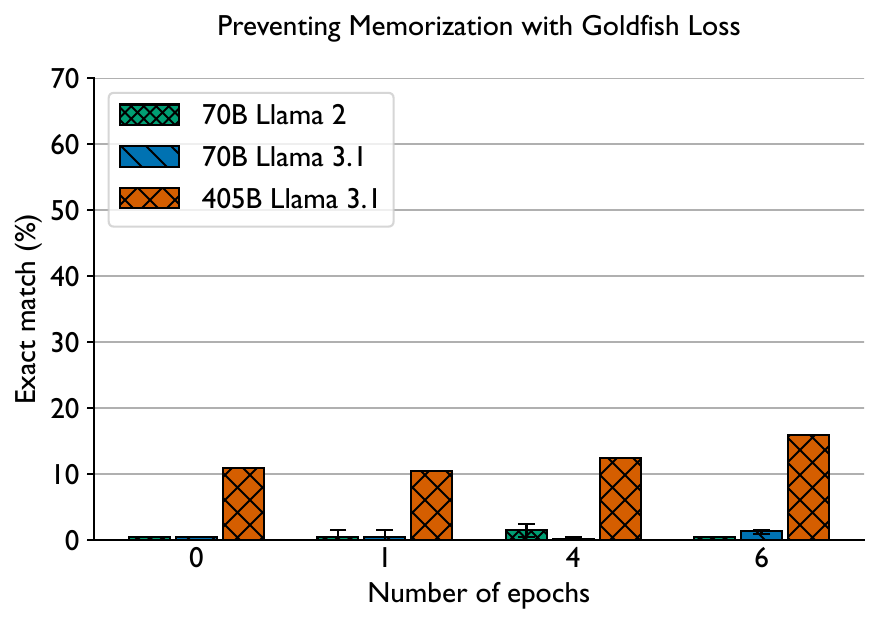}
    \caption{The impact of applying Goldfish Loss during training to mitigate
memorization in large models. The Exact Match rate reduces to levels comparable
to the control data.}
    \label{fig:goldfish}
\end{figure}

Figure~\ref{fig:goldfish} shows the results of re-running our training
experiments with Goldfish Loss activated (using Goldfish parameters k=2, h=13).
Even after continued training, memorization now reduces to levels comparable to the
control data (0 Ep). We do observe a small increase in memorization as the 405B
model trains, likely because the model has already memorized the masked tokens
from when it was pre-trained on Wikipedia. However, as we can see, the reduction in memorization when using the Goldfish Loss is significant, both for the 70B models and the 405B model.

\section*{Acknowledgment}
A large number of people helped us in securing access and allocations at
different supercomputing sites, and/or with improving the performance of our
framework, and we wish to thank each one of them. For compute time allocations
and reservations: Richard Gerber and Rebecca Hartman-Baker at NERSC/LBL;
Bronson Messer and Phil Roth at ORNL; Maria-Grazia Giuffreda at CSCS; and Jack
Wells at NVIDIA. For help with runs and software stack issues: Kevin Gott and
Peter Harrington at NERSC/LBL; Jens Glaser and Michael Sandoval at ORNL; Fabian
B\"osch, Theofilos Manitaras, Henrique Mendoni\c{c}a, and Fawzi Mohamed at
CSCS; Nicholas Malaya and Alessandro Fanfarillo at AMD; Tom Gibbs, and Josh
Romero at NVIDIA; Mark Stock and Mengshiou Wu at HPE; and others working behind
the scenes.

This material is based upon work supported in part by the ONR and OFOSR MURI
programs, and the National Science Foundation (IIS-2229885 \& IIS-2212182).  An
award for computer time was provided by the U.S.~Department of Energy’s (DOE)
Innovative and Novel Computational Impact on Theory and Experiment (INCITE)
Program. This research used resources of:~(1)~the Oak Ridge Leadership
Computing Facility at the Oak Ridge National Laboratory, which is supported by
the Office of Science of the U.S.~DOE under Contract
No.~DE-AC05-00OR22725,~(2)~the National Energy Research Scientific Computing
Center (NERSC), a DOE Office of Science User Facility using NERSC awards
DDR-ERCAP0029894 and DDR-ERCAP0029890, and~(3) the Swiss National
Supercomputing Centre (CSCS).

\IEEEtriggeratref{41}
\bibliography{./bib/pssg, ./bib/cite}
\bibliographystyle{IEEEtran}

\end{document}